\documentclass{article}

\usepackage[dblblindworkshop,final]{neurips_2026}
\workshoptitle{Interpretability for Discovery}

\usepackage[utf8]{inputenc}
\usepackage[T1]{fontenc}
\usepackage{hyperref}
\usepackage{url}
\usepackage{booktabs}
\usepackage{amsfonts}
\usepackage{amsmath}
\usepackage{amssymb}
\usepackage{nicefrac}
\usepackage{microtype}
\usepackage{xcolor}
\usepackage{graphicx}
\usepackage{multirow}
\graphicspath{{images2/}}

\title{Auditing Learned Dynamics in Weather Models} 

\author{%
  Skylar Gu \\
  Mila \\
  McGill University \\
  \texttt{linxuan.gu@mail.mcgill.ca} \\
  \And
  Christina Last \\
  Haziel \\
  \texttt{christina@haziel.co.uk} \\
}

\begin{document}
\maketitle

\begin{abstract}
Learned weather models are experimentally accessible computational systems whose internal representations can be inspected and manipulated. Using sparse autoencoder features from GraphCast, DeepMind's autoregressive weather forecasting model, we test whether causal relevance requires physical semantics, a question that matters for auditing whether AI forecasters used in high-assurance settings rely on reliable mechanisms. We also look at whether physically associated latent directions can be steered to improve extreme-event forecasts, where interventions are applied to the model's internal activations during the rollout. We find that ablating a feature locked to GraphCast's computational mesh increases global geopotential height error despite lacking conventional physical semantics. Second, suppressing convection- and low-level-vorticity-associated features weakens predicted tropical-cyclone deepening (the fall in central pressure during intensification), while amplifying them reduces intensity error, with amplification of convection features through feature scaling achieving a median of 51\% error reduction over seven storms. Together, these results separate interpretable, physical association from causal utility and establish sparse features as experimental coordinates for auditing and intervening on learned weather dynamics.
\end{abstract}

\section{Introduction}

Learned weather models now achieve forecast skill competitive with, and at times exceeding, numerical forecasting systems \citep{lam2023graphcast,bi2023pangu,price2025gencast}. Forecasts underpin high-assurance decisions such as insurance and natural-disaster response. As AI forecasts become more skillful, operational centres are beginning to adopt them. Thus, it's important to verify that these models learn reliable mechanisms for phenomena such as tropical cyclones. Mechanistic interpretability offers a way to audit their prediction mechanisms, but only if we know what its internal features do causally. Prior mechanistic interpretability work trained a sparse autoencoder (SAE) on GraphCast, a state-of-the-art machine-learning weather model, and identified model features associated with tropical cyclones, seasonal and diurnal cycles, and showed that intervening on a labelled tropical-cyclone feature produces coherent changes in evolving hurricanes \citep{macmillan2025}. However, these learned features may be overcomplete, architecture-dependent, and have spatially distributed, regime-dependent activations \citep{elhage2022superposition}. These patterns undermine their ability to serve as stable scientific variables. A feature may therefore be causally important without corresponding to a recognizable atmospheric quantity, while a physically labelled feature may improve a forecast when intervened on without acting as the physical process its label names. 

\subsection{Research questions}
\label{sec:rqs}
We investigate the following:
\begin{enumerate}
    \item Does a feature's causal relevance to forecast skill require a recognizable physical interpretation?
    \item Can features associated with physical processes (convection, low-level vorticity) provide causal leverage over a known forecast failure, the under-prediction of tropical-cyclone intensification?
\end{enumerate}

\subsection{Contributions}
\label{sec:contributions}
We make two distinctions about sparse representations in a learned weather model:
\begin{enumerate}
    \item \textbf{Causal relevance does not require physical semantics.} An architecture-specific feature locked to GraphCast's computational mesh can substantially affect forecast skill.
    \item \textbf{Physically associated features remain useful as causal intervention coordinates.} Suppressing convection- and low-level-vorticity-associated features weakens predicted tropical-cyclone intensification, while amplifying them can reduce intensity error. We report the outcome distribution across storms and compare results against firing-matched features and norm-matched random control vectors as well as statistical bias correction. 
\end{enumerate}
Together, these experiments separate physical interpretability from causal utility and identify when the two coincide.

\section{Methods}
\label{sec:methods}

We use the operational 0.25$^\circ$, 37-level GraphCast \citep{lam2023graphcast} and record the 512-dimensional mesh-node state after layer 8 (40{,}962 mesh nodes) at every 6-h analysis step from 2007 to 2018, teacher-forced using ERA5 data (that is, each step's input is the ERA5 state rather than the model's own previous output). ERA5 is a reanalysis, not a forecast: it combines a model with historic observations to estimate the atmospheric state of the past \citep{hersbach2020era5}. For feature analysis, we use the pretrained layer-8 SAE from \citet{macmillan2025}.

For each SAE feature $f$ and ERA5 field $m$, we compute $A_{f,m}$, the mean standardized anomaly of $m$ over nodes and times where $f$ fires, using 24 independent reanalysis windows. For a fixed feature $f$, its physical associations are described by the vector:

\[
\mathbf{A}_f
=
\bigl(
A_{f,m_1},\ldots,A_{f,m_8}
\bigr).
\]

Physical concept labels are assigned by measuring features' associations with the eight standard ERA5 physical variables by calculating the mean standardized anomaly of each variable where the feature fires (see Appendix~\ref{app:labels} for full details). A feature labelled ``ascent,'' for example, fires at nodes where mid-tropospheric ascent is unusually strong relative to its global distribution. These labels are assigned by us from statistical association with atmospheric conditions; they do not characterise where ascent occurs in a physical model, and do not imply that a feature activation is an estimator of the corresponding ERA5 field.

We call a feature grid-locked when its activation support stays attached to GraphCast mesh nodes under longitude rotations of the atmospheric input (shifting the input fields east-west by a fixed angle, e.g.\ 90$^\circ$, so every weather pattern moves to a different longitude), rather than moving with the meteorological state. We contrast grid-locked features, whose activation is tied to fixed mesh locations rather than weather, with state-dependent, physically labelled features, and measure the effect of ablating each on global $z_{500}$ forecast RMSE at $+48$\,h.

All interventions are applied to the layer-8 activations during the rollout: at each step we modify the selected SAE feature activations, decode, and let GraphCast continue autoregressively, so later steps respond to the change. For each storm we (1) perturb the activations of the features associated with deepening, (2) re-run GraphCast from the same ERA5 initial state, (3) compare the result with the unperturbed GraphCast forecast, and (4) compare both with ERA5.

To test for causal effect on forecast, we intervene on feature groups associated with hurricane formation. We look at seven rapidly intensifying tropical cyclones: Ida 2021, Michael 2018, Haishen 2020, Goni 2020, Haiyan 2013, Patricia 2015, and Wilma 2005. Each is rolled 96\,h (16 steps) from the ERA5 analysis preceding intensification; the intervention in Eq.~\ref{eq:intervention} is applied at every step to mesh nodes, by scaling their activation above a storm-free reference level, within 1{,}500\,km of the storm centre at initialization, with radius chosen to generously encompass regions of convective anomaly (as the effect saturates by 750\,km). The reference level, a feature's typical activation on storm-free days at the same location and time of year, is denoted by $f_{\mathrm{normal}}$.

\begin{equation}
f' = f + (\alpha-1)\max(f-f_{\mathrm{normal}},0),
\label{eq:intervention}
\end{equation}

Here, the scaling factor \(\alpha\) sets the intervention strength: \(\alpha=1\) leaves the feature unchanged, \(\alpha=0\) removes activation above its normal level, and \(\alpha>1\) amplifies it. The intervention is re-applied every 6\,h throughout the forecast, analogous to repeated activation steering during autoregressive generation \citep{rimsky2024caa}.


We use two kinds of controls which receive the same operator: random directions whose per-node magnitude equals that of the given intervention and SAE features matched to the target on global firing rate (within 15\%). The random directions thus isolate the effect of the direction of the intervention. However, they can push activations off distribution \citep{zhang2024patching}, so we also use firing-rate-matched features: real SAE features that fire about as often as the target feature but have no physical association with the storm. Because they are genuine in-distribution features, they test whether any feature of similar activity gives the same effect, which isolates the contribution of the target feature's content. Full intervention and control details are given in Appendix~\ref{app:interventions}. Code, pre-registration documents, and scripts are available at \url{https://github.com/Skylar-gu/causal-interp-climate}. All weights and datasets used are publicly released and there is no model retraining.

\section{Results}
\label{sec:results}

\subsection{Architecture-specific representations can be causally relevant}
\label{sec:results-gridlocked}

We first ask whether features that contribute to forecast skill necessarily correspond to recognizable atmospheric structure. \citet{macmillan2025} found that GraphCast contains features whose activations are locked to its computational mesh, referring to them as ``grid-locked'' in contrast to content features which shift geographically over time. These do not correspond to a recognizable physical variable, but often do exhibit a meaningful pattern. We define grid-locked features as those that stay on the mesh under longitude rotations of the input fields (e.g.\ wind and temperature). 

\begin{figure}[t]
    \centering
    \includegraphics[width=\linewidth]{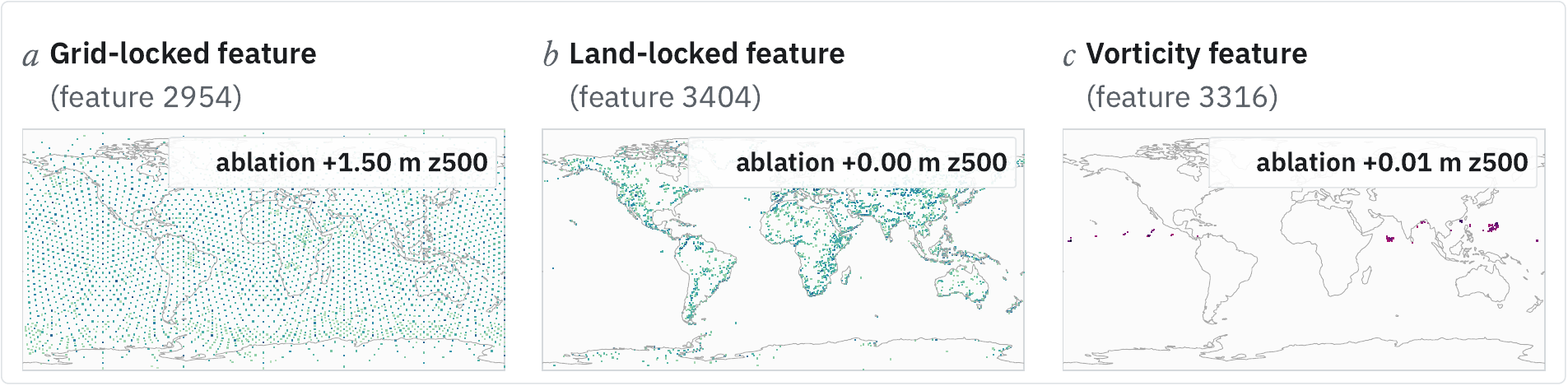}
    \caption{\textbf{Causal relevance does not require physical semantics.} Mean activation footprints on the GraphCast mesh for (a) a grid-locked feature, (b) a land-locked feature with fixed pattern and comparable coverage, and (c) a vorticity-associated feature. Ablation scores show the change in global $z_{500}$ RMSE, averaged over six initial conditions.}
    \label{fig:gridlocked}
\end{figure} 

We test these features for causal relevance by measuring global $z_{500}$, geopotential height at the 500 hPa pressure level. We compare the change in $z_{500}$ RMSE (m) under ablation across features and find that forecast importance can arise without physical interpretability. Figure~\ref{fig:gridlocked} shows one such grid-locked feature with two non-grid-locked features. This grid-locked feature fires at the same mesh nodes regardless of the atmospheric state. Ablating it raises $+48$\,h global geopotential height ($z_{500}$) error by 1.50\,m (19\% of the baseline error), whereas ablating a feature of comparable coverage (Figure~\ref{fig:gridlocked}b) results in almost no additional error. A feature corresponding to vorticity also results in a low global error but has a strong local effect (Figure~\ref{fig:conv}a). We note that these effects are not uniform across ERA5 fields and many grid-locked features, especially those with smaller footprints, have smaller effects on forecast accuracy when ablated (Appendix~\ref{app:gridlock}, Figure~\ref{fig:gridlockedlong}).

These results reveal that architecture-specific representations can become causally load-bearing without conventional physical interpretations, instead providing model-specific computational affordances. The model may exploit coordinates or shortcuts meaningful within its architecture rather than the ontology of a numerical weather model. Physical plausibility is therefore not a prerequisite for causal relevance. One possible reading is that the grid-locked features act as a channel through which local, physically labelled features influence skill across the globe, since mesh nodes are how information moves between regions. We test this only preliminarily (see Section~\ref{sec:future work}). We next turn to whether features that do have a physical association can provide causal leverage over a forecast failure. These interventions use state-dependent, physically labelled features, not grid-locked ones.

\subsection{Direct interventions reveal selective causal leverage}
\label{sec:results-interventions}

We look at tropical-cyclone intensification as a test case. GraphCast systematically under-forecasts rapid tropical-cyclone intensity, i.e.\ the deepening of rapidly intensifying storms \citep{demaria2025tc}. Deep convection and low-level cyclonic circulation (vorticity) are central to tropical-cyclone intensification. Having identified SAE features associated with observable signatures of these processes, we therefore ask whether suppressing those features inhibits predicted deepening, and whether amplifying them can reduce forecast error.

\begin{figure}[t]
    \centering
    \includegraphics[width=\linewidth]{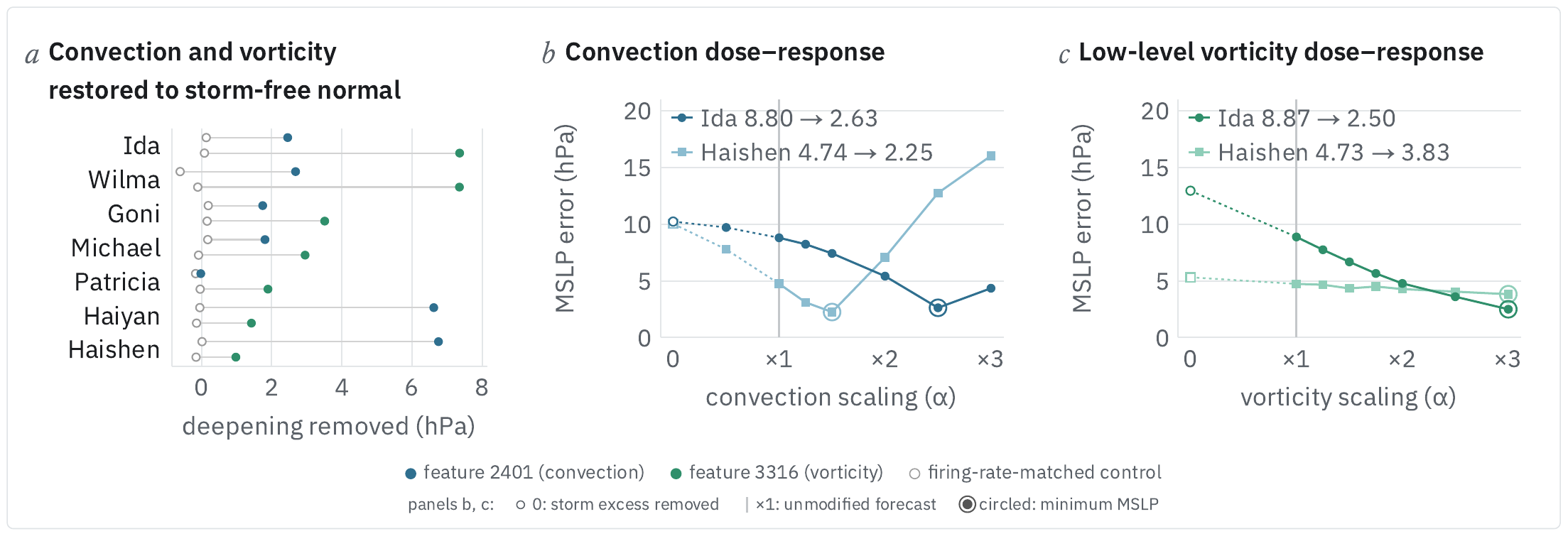}
    \caption{\textbf{Removing convection or vorticity features weakens forecast storms; amplifying them can correct under-predicted ones.} (a) Deepening removed when the single convection (2401) and low-level vorticity (3316) features are restored to storm-free normal ($\alpha=0$), with firing-rate-matched controls. (b), (c) Legend values give MSLP error for the unmodified forecast and at its minimum (circled). The optimal feature scaling is storm-dependent: $\alpha=2.5$ for Ida and $\alpha=1.5$ for Haishen for convection, beyond which error increases; $\alpha=3$ for both for vorticity, though Haishen barely moves, as the feature fires weakly in the target region.}
    \label{fig:convection}\label{fig:conv}\label{fig:conv_vort}
\end{figure}

\begin{figure}
    \centering
    \includegraphics[width=1\linewidth]{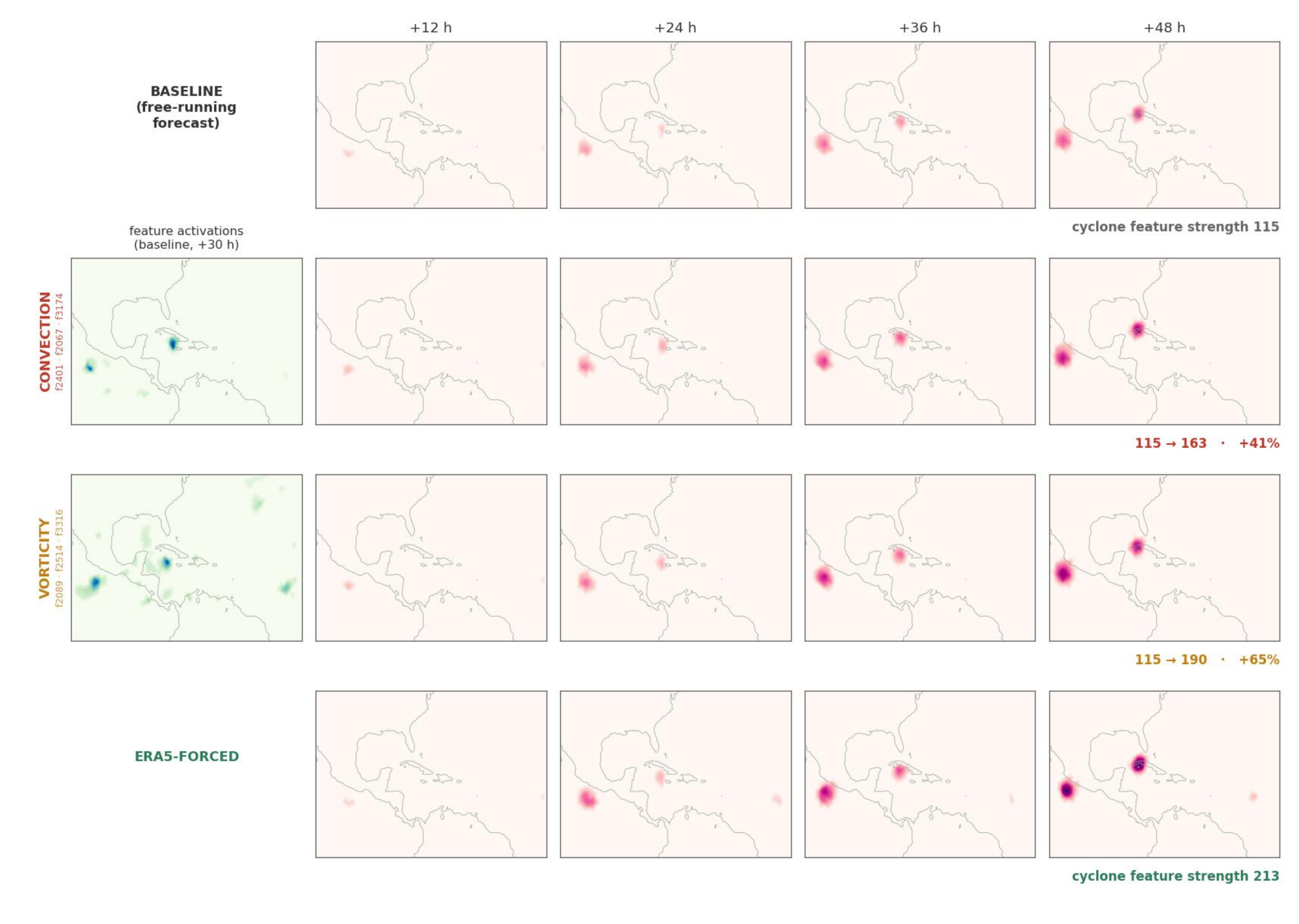}
    \caption{\textbf{Effect of interventions on the model's tropical-cyclone representation.} Hurricane Ida (2021). The cyclone feature strength is the total activation of the TC feature summed over the shown Gulf/Caribbean box. Maps show activation of the previously identified TC feature during the baseline rollout, after doubling convection- or vorticity-associated features at every forecast step, and under ERA5 teacher forcing. Columns to the right show successive lead times (+12, +24, +36, +48\,h) of one 6-hourly rollout with the intervention applied at every step, so they show evolution through the rollout, not increasing intervention strength. The left column shows where the two feature groups fire in the baseline run at +30\,h. By $+48$\,h, TC-feature strength increases from the baseline to 163 under convection amplification and 190 under vorticity amplification, toward the score of 213 under ERA5 forcing.}
    \label{fig:interventions} 
\end{figure}

We run two kinds of intervention with the operator in Eq.~\ref{eq:intervention}. First, we restore features to their storm-free normal and measure how much of the forecast storm's deepening is removed. Second, we amplify them and score the forecast by MSLP error, the RMSE of the storm's minimum sea-level pressure against ERA5 over the intensification window. Each intervention is applied to the single features corresponding to convection and low-level vorticity on all seven storms (Figure~\ref{fig:conv}). We then examine the effects of amplifying groups of three features for convection and vorticity on the activation values of the tropical cyclone feature (Figure~\ref{fig:interventions}).

Restoring either feature toward its quiet-state activations reduces forecast storm deepening much more than firing-rate-matched features (Figure~\ref{fig:conv}a). Across seven storms, restoring the convection feature removes a median 2.5\,hPa of deepening and the low-level-vorticity feature a median 3.0\,hPa, against a median of less than 0.1\,hPa for the firing-rate-matched features.

While removing these features' storm-driven excess can degrade cyclone forecasts, amplifying them can improve forecasts (Figure~\ref{fig:conv}b,c). We score forecasts by MSLP error, the RMSE of the storm's minimum sea-level pressure against ERA5 over the intensification window. Scaling the convection feature at optimal $\alpha$ values results in a median improvement of 51\% in MSLP error. One storm experienced no effect across the sweep of $\alpha$ values and the other two storms showed improvements at lower $\alpha$ values but not higher ones. Amplifying the low-level-vorticity feature lowers MSLP error steadily as the scaling factor increases, without over-deepening any storm: at $\alpha=3$ all seven storms had a median improvement of 42\%. See Table~\ref{tab:single} for details. This response pattern shows that the intervention changes the predicted storm, not only latent activity, but the best scaling factor is storm-dependent. The dose-response curves therefore identify state-dependent causal levers rather than universal corrections.

\begin{figure}[t]
    \centering
    \includegraphics[width=\linewidth]{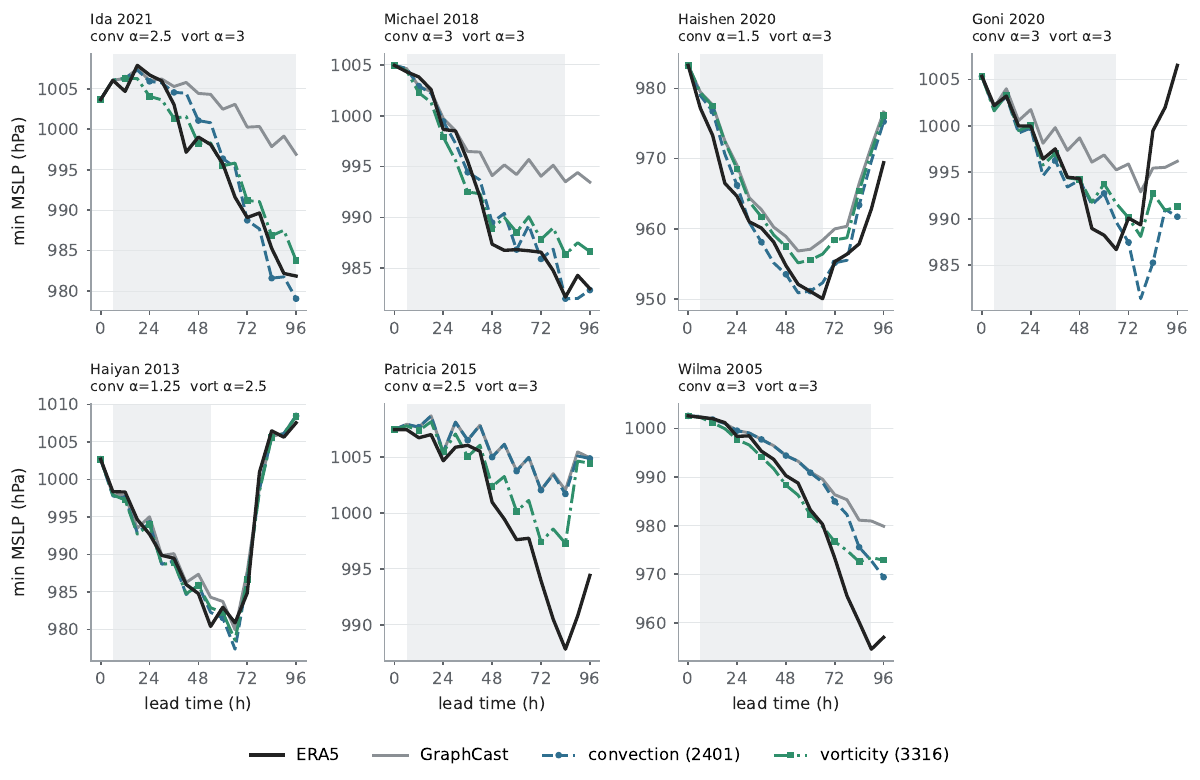}
    \caption{\textbf{Minimum sea-level pressure against lead time, all seven storms.} ERA5, the unmodified GraphCast forecast, and features 2401 and 3316 amplified at each storm's best scaling factor (Table~\ref{tab:single}). Re-rolled MSLP errors reproduce Table~\ref{tab:single} within 0.5 hPa for all storms and features.}
    \label{fig:mslp_time}
\end{figure}

Looking at the minimum  sea-level pressure across lead time (Figure~\ref{fig:mslp_time}), we see the same pattern across storms. The unperturbed forecast tracks ERA5 closely over the first half of the 96-h window and falls increasingly behind as the storm intensifies, and amplification at the best scaling factor narrows this gap. Neither feature is uniformly better: convection gives the lower error in Michael, Haishen, Goni and Haiyan, vorticity in Ida, Patricia and Wilma, with Patricia and Wilma having relatively smaller gains.


Consistently, doubling the convection or vorticity feature groups increases the activation values of the tropical-cyclone feature as identified by \citet{macmillan2025}. The values increase up to 65\%, making up most of the 85\% increase needed to reach the feature's activation values under ERA5 teacher forcing (Figure~\ref{fig:interventions}). This result links the feature interventions' increase of storm deepening to intensification of its internal representation of cyclones towards the direction corresponding to decreased forecasting error.

We also compare amplification with bias correction, the standard statistical post-processing step for removing systematic forecast error \citep{glahn1972mos,cui2012bias}: for each storm, we average the ERA5-minus-forecast difference in minimum MSLP at each 6-h lead over the other six storms and add it to that storm's unmodified forecast. With the scaling factor chosen leave-one-out, it lowers MSLP error to a median 2.3\,hPa, against 7.4\,hPa for the convection feature and 2.5\,hPa for the vorticity feature (Table~\ref{tab:offset}). However, despite not always being the more effective correction, amplification acts on the model's internal state so the whole forecast storm evolves consistently with the change, and locates the intensity error in specific, interpretable features. 


\section{Discussion}
\label{sec:discussion}

Taken together, the experiments separate physical interpretability from causal relevance. Figure~\ref{fig:gridlocked} establishes causal relevance without conventional physical semantics, while Figures~\ref{fig:conv} and~\ref{fig:interventions} show that physically associated features can nevertheless provide state-dependent causal leverage over a specific forecast failure. However, these features need not constitute regime-independent mechanisms as individual features may participate in distributed or superposed representations rather than correspond one-to-one with underlying variables \citep{cunningham2023, bricken2023}. SAE features may therefore be most useful as coordinates of model state whose causal effects depend on the surrounding configuration rather than solely on fixed feature semantics.

Our interventions identify physically associated internal directions whose manipulation can address systematic failure modes retained by the model, including under-prediction of tropical-cyclone intensity \citep{demaria2025tc}. Suppressing convection- and low-level-vorticity-associated features reduces predicted intensification, and amplifying them can substantially reduce intensity error, although the response depends on both storm state and intervention strength. Internal interventions thus provide an alternative route to post-hoc model correction as these adjustments adjust the trajectory of the model, with future rollouts continuing on a hopefully more aligned path.


\section{Limitations}
\label{sec:limitations}

Our physical labels describe statistical associations between feature activation and ERA5 fields. They do not establish that an SAE feature represents that physical variable or mechanism. The tropical-cyclone intervention study covers only seven rapidly intensifying storms plus five in the Southern Hemisphere for convection (Table~\ref{tab:sh}), and the dose-response varies substantially across storms, meaning no $\alpha$ value is optimal across storms. With $\alpha$ chosen leave-one-out instead of using the optimal $\alpha$ for each storm, convection amplification no longer beats the unmodified forecast on median MSLP error (7.43 vs.\ 6.40\,hPa; Table~\ref{tab:single}).

We note that model improvements are measured against ERA5, the training target, whose minimum pressure is more than 40\,hPa shallower than the best measurements for the seven storms due to the coarse resolution of the reanalysis. Agreement with it thus measures closeness to an observation-informed state rather than forecast skill achievable in real time. We note also that interventions establish causal leverage within GraphCast's computation, but do not imply the perturbed representations are in closer agreement with the true past observations.

\section{Future Work}
\label{sec:future work}

\paragraph{Evaluation.} We score minimum-pressure RMSE over the intensification window against ERA5. Future work could (i) score against best-track observations, since ERA5 underestimates storm depth; and (ii) test whether baseline feature activity in the storm region predicts the optimal $\alpha$, which would turn the storm-dependent dose-response into a usable correction rather than a hindsight-tuned sweep.

\paragraph{Mechanism and breadth.} If grid-locked features carry local signals outwards, jointly ablating one with a convection or vorticity intervention should weaken that intervention's effect at locations away from the storm. In a preliminary test, ablating one of the grid-locked features (feature 2954) had negligible effects on the changes produced by amplifying the feature associated with vorticity.

\paragraph{Activation-space correction.} We also tried injecting the mean layer-8 activation difference between ERA5 teacher forcing and the free rollout, averaged over the other six storms, at every disk node and rollout step. It lowered MSLP error in 5 of 7 storms, but using each storm's own difference improved only 4 of 7, so the test is inconclusive and we leave it to future work to untangle the reasons behind this result.

More storms, basins, hazards (e.g.\ extratropical cyclones, atmospheric rivers, heatwaves) and forecasting models should be tested to show whether such features recur across learned forecasters, as needed for them to serve as audit diagnostics in high-assurance use.

\section{Conclusion}
\label{sec:conclusion}

In this paper, we evaluate the physical plausibility and causal relevance of sparse features in a large learned weather model. Features tied to GraphCast's computational mesh can be causally important for forecast skill despite lacking conventional physical semantics, while features associated with atmospheric convection and low-level vorticity provide causal leverage over tropical-cyclone intensification. MSLP error measured against ERA5 was reduced by 2--75\% across the seven storms by amplifying the feature correlated with convection, and 15--71\% by intervening on the vorticity feature, with medians of 51\% and 42\% respectively at each storm's best scaling factor. Applying these interventions results in closer forecasted and ERA5 MSLP values across lead time (Figure~\ref{fig:mslp_time}). Sparse internal representations can thus provide useful coordinates for extreme events. 

These results separate physical interpretability from causal utility while showing that the two can coincide. For machine-learning weather and climate models, SAEs provide experimental coordinates for identifying, perturbing, and tracking internal representations of atmospheric phenomena. Beyond model audit and forecast correction, such representations could support scientific discovery by revealing recurrent internal structure around extreme events, and suggesting candidate relationships or precursors that can then be tested against observations and physical theory. 

\section{Responsible use}
\label{sec:responsible-use}

We present interventions as tools for scientific audit and model understanding, while noting that these mechanisms can be used to distort operational forecasts and affect downstream decision-making.

\begin{ack}
This work was funded by the Compound Reverie Grant. Compute was provided by World Mechanics and the Digital Research Alliance of Canada. The authors declare no competing interests.
\end{ack}


\bibliographystyle{plainnat}
\bibliography{refs}

\appendix

\section{Feature labeling and calibration}
\label{app:labels}

The association vector $\mathbf{A}_f$ (Section~\ref{sec:methods}) is computed over 24 independent reanalysis windows, and a feature's candidate label is the ERA5 field with the largest $|A_{f,m}|$. Labels are calibrated against 127 rigid longitude rotations that preserve feature footprint size, shape, and latitude while breaking alignment with the meteorological field. We compute empirical $p$-values from this null and apply Benjamini--Hochberg FDR correction at $q=0.05$ \citep{benjamini1995fdr}. To avoid labelling features similarly associated with several fields, the strongest association must also be separated from the second strongest by more than 95\% of the corresponding separations under the rotation null. On a null dataset with no true feature--field alignment, the procedure labels 2.7\% of features.

The calibrated labels used for the interventions, including the rotation null and the 2.7\% rate above, use globally standardized associations with four variables: 850-hPa relative vorticity, 600-hPa specific humidity, mid-tropospheric ascent, and deep-layer shear.

\section{Intervention protocol and controls}
\label{app:interventions}

\paragraph{Intervention groups.}
The convection/ascent group comprises features 2401, 2067, and 3174, while the low-level-spin group comprises 2089, 2514, and 3316. Feature 3243, the previously labelled tropical-cyclone feature, is excluded from both groups. Figure~\ref{fig:conv} intervenes on the single features 2401 and 3316 and Figure~\ref{fig:interventions} on both groups.

\paragraph{Reference level and controls.}
For each feature and storm, $f_{\mathrm{normal}}$ is estimated from storm-free analogue dates at the same location and time of year. Analogues containing substantial tropical-cyclone feature activation are excluded. If the feature does not fire on any surviving analogue, $f_{\mathrm{normal}}=0$.

Both controls receive the identical operator, spatial mask and scaling factor. The firing-rate features are features 3667, 2875 and 2850, matched to the convection features' global firing rate within 15\% and restricted to tropical centroids, drawn with a fixed seed and frozen before any results. Inside the storm they fire at about 2\% of the convection group's level. The same three features serve as the control for feature 3316. 

The random-direction controls are five Gaussian unit vectors in the 512-dimensional layer-8 activation space, projected orthogonal to the span of the convection decoder vectors, each applied with the same per-node magnitude as the target's intervention at that scaling factor; they were run for the convection group only, so Figure~\ref{fig:conv} includes only the locality check. 

\paragraph{Intervention mask.}
The intervention is applied within a fixed 1{,}500\,km disk centred on the storm at initialization and is re-applied at every forecast step. The mask is not recentered during the rollout.

\paragraph{Single-feature sweeps.} Table~\ref{tab:single} gives the amplification sweeps of the single features on all seven storms. Feature 3316 lowers MSLP error steadily as $\alpha$ rises and does not over-deepen any storm. Amplification of feature 2401 produces varying error improvements with highly fluctuating scaling factor optima. 

\begin{table}[h]
\centering
\small
\caption{Minimum MSLP error (hPa) against scaling factor $\alpha$ for each feature. Parentheses give the percentage change in MSLP error relative to the feature's unmodified forecast (\%; negative is better).}
\label{tab:single}
\setlength{\tabcolsep}{4pt}
\resizebox{\textwidth}{!}{%
\begin{tabular}{lccccccc}
\toprule
$\alpha$ & Ida & Michael & Haishen & Goni & Haiyan & Patricia & Wilma \\
\midrule
\multicolumn{8}{l}{\footnotesize\emph{Convection, feature 2401}} \\[3pt]
1    & 8.87 & 6.40 & 4.87 & 4.71 & 1.81 & 6.56 & 11.37 \\
1.25 & 8.21 {\scriptsize($-7$)} & 6.05 {\scriptsize($-5$)} & 3.08 {\scriptsize($-37$)} & 4.45 {\scriptsize($-6$)} & \textbf{1.23} {\scriptsize($-32$)} & 6.54 {\scriptsize($0$)} & 11.14 {\scriptsize($-2$)} \\
1.5  & 7.43 {\scriptsize($-16$)} & 5.72 {\scriptsize($-11$)} & \textbf{2.37} {\scriptsize($-51$)} & 4.30 {\scriptsize($-9$)} & 1.67 {\scriptsize($-8$)} & 6.53 {\scriptsize($0$)} & 10.71 {\scriptsize($-6$)} \\
2    & 5.25 {\scriptsize($-41$)} & 4.67 {\scriptsize($-27$)} & 7.02 {\scriptsize($+44$)} & 3.73 {\scriptsize($-21$)} & 4.51 {\scriptsize($+149$)} & 6.50 {\scriptsize($-1$)} & 10.24 {\scriptsize($-10$)} \\
2.5  & \textbf{2.59} {\scriptsize($-71$)} & 3.36 {\scriptsize($-48$)} & 12.88 {\scriptsize($+164$)} & 2.90 {\scriptsize($-38$)} & 7.54 {\scriptsize($+317$)} & \textbf{6.44} {\scriptsize($-2$)} & 9.69 {\scriptsize($-15$)} \\
3    & 4.06 {\scriptsize($-54$)} & \textbf{1.62} {\scriptsize($-75$)} & 16.10 {\scriptsize($+231$)} & \textbf{1.98} {\scriptsize($-58$)} & 9.58 {\scriptsize($+429$)} & 6.51 {\scriptsize($-1$)} & \textbf{8.94} {\scriptsize($-21$)} \\
\midrule
\multicolumn{8}{l}{\footnotesize\emph{Low-level vorticity, feature 3316}} \\[3pt]
1    & 8.89 & 6.51 & 4.59 & 4.63 & 1.79 & 6.53 & 11.19 \\
1.25 & 7.79 {\scriptsize($-12$)} & 5.90 {\scriptsize($-9$)} & 4.70 {\scriptsize($+2$)} & 4.35 {\scriptsize($-6$)} & 1.64 {\scriptsize($-8$)} & 6.22 {\scriptsize($-5$)} & 10.24 {\scriptsize($-8$)} \\
1.5  & 6.68 {\scriptsize($-25$)} & 5.25 {\scriptsize($-19$)} & 4.51 {\scriptsize($-2$)} & 3.98 {\scriptsize($-14$)} & 1.55 {\scriptsize($-13$)} & 5.84 {\scriptsize($-11$)} & 9.33 {\scriptsize($-17$)} \\
2    & 4.78 {\scriptsize($-46$)} & 4.25 {\scriptsize($-35$)} & 4.34 {\scriptsize($-5$)} & 3.40 {\scriptsize($-27$)} & 1.44 {\scriptsize($-20$)} & 5.12 {\scriptsize($-22$)} & 8.10 {\scriptsize($-28$)} \\
2.5  & 3.52 {\scriptsize($-60$)} & 3.23 {\scriptsize($-50$)} & 3.95 {\scriptsize($-14$)} & 2.83 {\scriptsize($-39$)} & \textbf{1.39} {\scriptsize($-22$)} & 4.43 {\scriptsize($-32$)} & 7.08 {\scriptsize($-37$)} \\
3    & \textbf{2.54} {\scriptsize($-71$)} & \textbf{2.50} {\scriptsize($-62$)} & \textbf{3.90} {\scriptsize($-15$)} & \textbf{2.40} {\scriptsize($-48$)} & 1.47 {\scriptsize($-18$)} & \textbf{3.81} {\scriptsize($-42$)} & \textbf{6.74} {\scriptsize($-40$)} \\
\bottomrule
\end{tabular}}
\end{table}

\begin{table}[h]
\centering
\small
\caption{Convection group amplified at $\alpha=2$ on five Southern Hemisphere storms. Negative change means lower MSLP error.}
\label{tab:sh}
\begin{tabular}{lcccc}
\toprule
Storm & Unmodified (hPa) & $\alpha=2$ (hPa) & Change (hPa) & Change (\%) \\
\midrule
Marcus 2018   & 9.13 & 2.76 & $-6.37$ & $-70$ \\
Veronica 2019 & 4.80 & 2.96 & $-1.84$ & $-38$ \\
Fantala 2016  & 5.95 & 4.16 & $-1.79$ & $-30$ \\
Harold 2020   & 7.41 & 7.06 & $-0.36$ & $-5$  \\
Winston 2016  & 8.79 & 9.60 & $+0.81$ & $+9$  \\
\bottomrule
\end{tabular}
\end{table}

\paragraph{Bias-correction baseline.} 

As a baseline for forecast skill we use the simplest standard post-processing \citep{glahn1972mos,cui2012bias}: for each storm, add the mean ERA5-minus-forecast minimum-MSLP difference at each 6-h lead, averaged over the other six storms. It has the lowest error in four of seven storms (Table~\ref{tab:offset}). It beats the convection feature in all seven, most clearly in Haishen and Haiyan, which convection amplification overshoots, while the vorticity feature has the lowest error in Ida, Haiyan and Wilma. The comparison favours the correction, which is fitted only on rapidly intensifying storms and changes only the reported pressure, not the model's state. 

\begin{table}[h]
\centering
\small
\caption{MSLP error (hPa) with the scaling factor chosen leave-one-out: for each storm, $\alpha$ is the scaling factor with the lowest median error over the other six storms, and the bias correction is likewise fitted on the other six. Convection is feature 2401 and vorticity feature 3316, each scaled independently (Table~\ref{tab:single}); the selected scaling factor is $\alpha=3$ for 3316 in every storm and varies between 1.5 and 2.5 for 2401. Bold marks the lowest error for each storm.}
\label{tab:offset}
\setlength{\tabcolsep}{4pt}
\begin{tabular}{lcccccccc}
\toprule
 & Ida & Michael & Haishen & Goni & Haiyan & Patricia & Wilma & Median \\
\midrule
Convection     & 7.43 & 5.72 & 12.88 & 3.73 & 7.54 & 6.50 & 9.69 & 7.43 \\
Vorticity      & \textbf{2.54} & 2.50 & 3.90 & 2.40 & \textbf{1.47} & 3.81 & \textbf{6.74} & 2.54 \\
Bias-corrected & 3.96 & \textbf{1.78} & \textbf{2.70} & \textbf{1.05} & 1.98 & \textbf{2.28} & 7.25 & \textbf{2.28} \\
\bottomrule
\end{tabular}
\end{table}

\section{Details on grid-locked features}
\label{app:gridlock}

Grid-locked feature ablations are evaluated over six approximately two-month-spaced initial conditions from 2020, outside GraphCast's training period. Each feature, or its matched control, is set to zero throughout the +48\,h rollout.

\paragraph{Grid-lock criterion.}
We rotate the atmospheric input by $45^\circ$, $90^\circ$, $135^\circ$, $180^\circ$, and $270^\circ$ in longitude and compare each feature's activation after rotation with its unrotated activation at (i) the same mesh nodes and (ii) the geographically rotated nodes. We define
\begin{equation}
    s_f =
    \operatorname{corr}(\text{rotated},\text{stays})
    -
    \operatorname{corr}(\text{rotated},\text{follows}).
\end{equation}
Positive $s_f$ indicates that the feature remains attached to the computational mesh rather than following the atmospheric state. Features with self-similar patterns under a given rotation are excluded for that angle. We classify a feature as grid-locked when $s_f \geq 0.5$ for all five rotations; six features satisfy this criterion.

\begin{figure}
    \centering
    \includegraphics[width=0.9\linewidth]{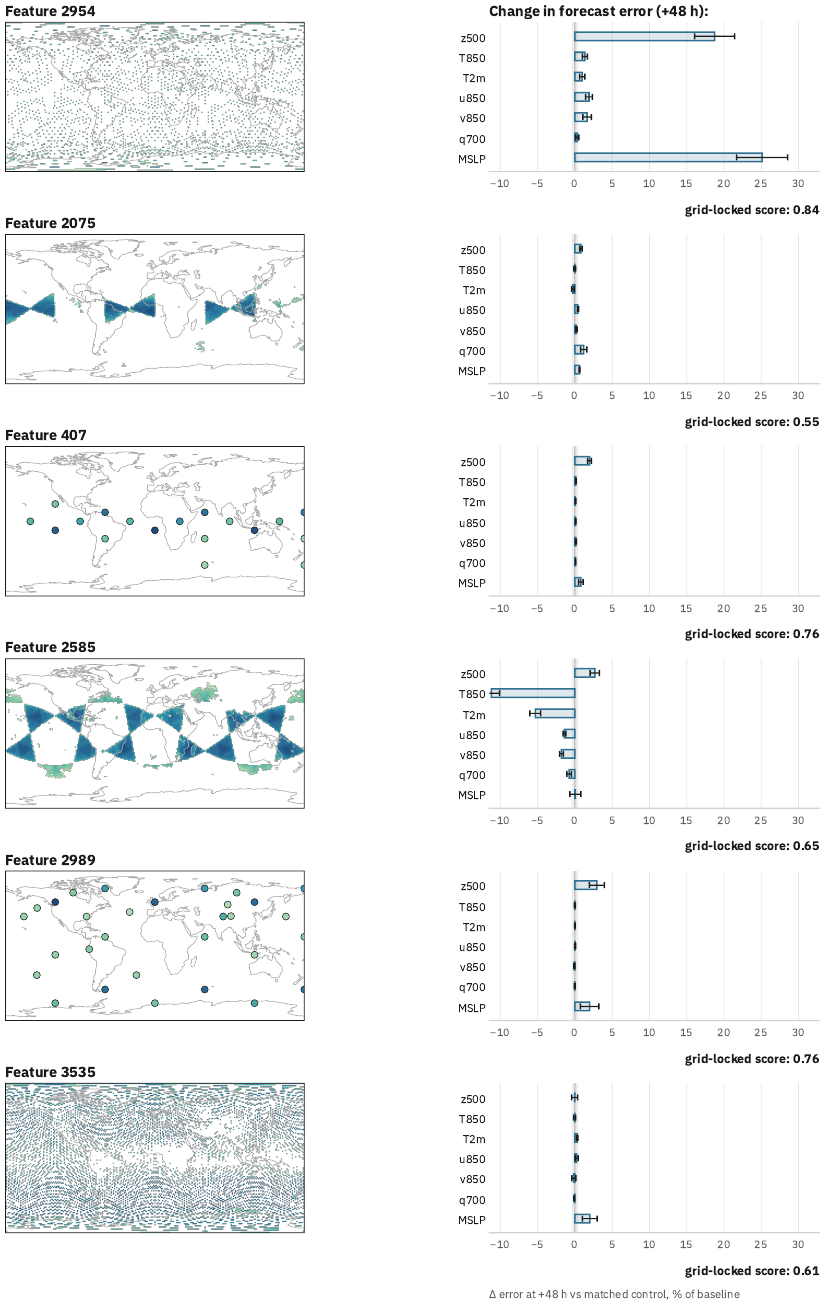}
    \caption{\textbf{Forecast effects of grid-locked features.} Mean activation footprints and change in global RMSE against ERA5 at +48\,h after feature removal.}
    \label{fig:gridlockedlong}
\end{figure}

\begin{figure}
    \centering
    \includegraphics[width=0.9\linewidth]{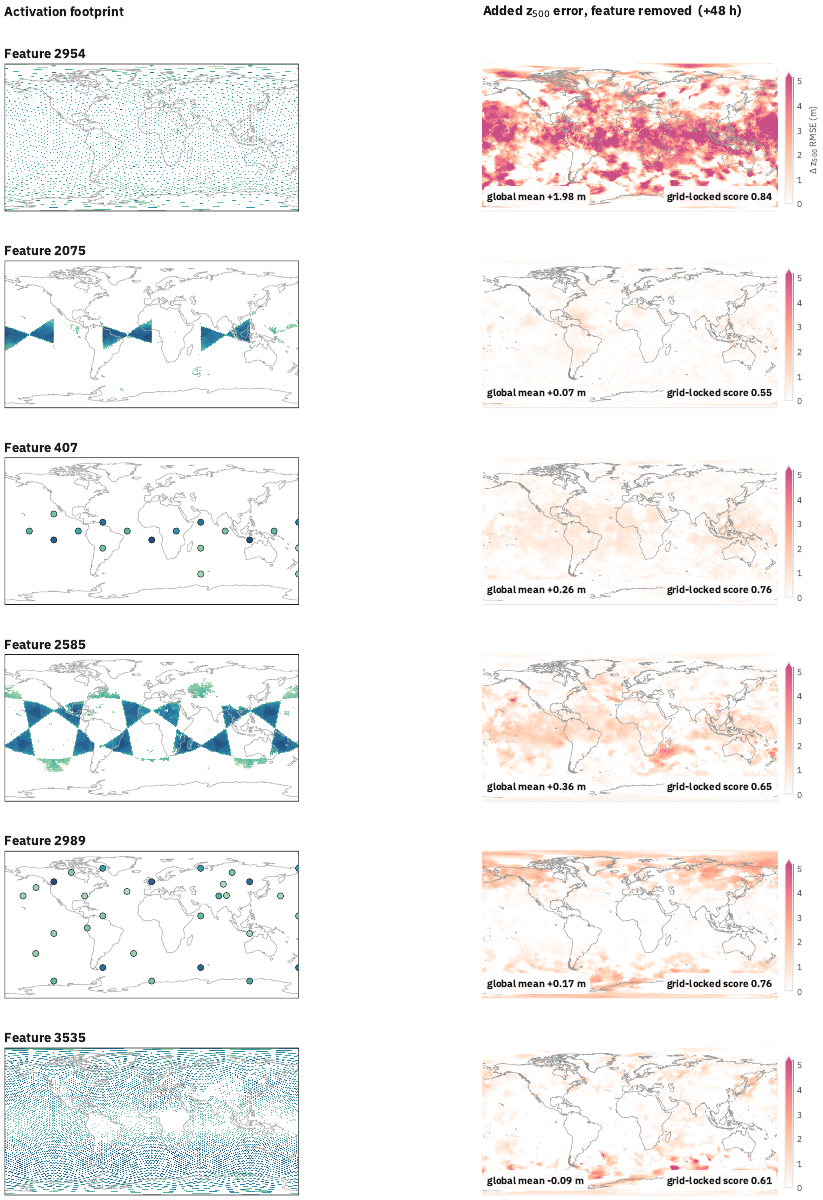}
    \caption{\textbf{Spatial effects of grid-locked feature removal.} Activation footprints and corresponding change in $z_{500}$ RMSE at +48\,h.}
    \label{fig:gridlockedz500}
\end{figure}

Figure~\ref{fig:gridlockedlong} summarizes the global, field-wise effects of these ablations, while Figure~\ref{fig:gridlockedz500} shows the corresponding spatial distribution of \(z_{500}\) error.

\paragraph{Ablation controls.}
Each grid-locked feature is paired with a non-grid-locked SAE feature matched on spatial coverage and mesh connectivity. We report the difference in forecast error between ablating the grid-locked feature and ablating its matched control:
\begin{equation}
    d_i =
    \mathrm{RMSE}(\text{feature ablated})_i -
    \mathrm{RMSE}(\text{control ablated})_i.
\end{equation}

\end{document}